\documentclass{article}
\usepackage{latexsym}
\usepackage{graphicx}

\usepackage{makeidx}  % allows for indexgeneration
\usepackage{eurosym}
\usepackage{amsfonts}

\usepackage{epsfig}
\usepackage{color}
\usepackage{amsfonts}

\usepackage{amsmath}
\begin{document}

% \singlespacing

\title{Relational Models}

\author{Volker Tresp$^{1}$   % same address
and Maximilian Nickel$^{2}$ \\
$^{1}$Siemens AG and Ludwig Maximilian University of Munich \\
{$^{2}$Facebook AI Research} \\
Draft \\
 To appear in the 2nd edition of the \\ \emph{Encyclopedia of Social Network Analysis and Mining}, Springer.
          % R.Roe works at addr. no. 2,
 }          % Santa Claus at addr. no. 3

%\institute{Siemens AG\\    % first address
%Otto-Hahn-Ring 6 \\
%81739 M\"unchen, Germany
%\and                               % second address
%Ludwig Maximilian University of Munich\\
%Department of Computer Science \\
%Oettingenstra{\ss}e 67 \\
% 80538 M\"unchen, Germany \\
%Phone: +49 89 636-49408\\
%Fax:   +49 89 636-42284 \\
%E-mail: volker.tresp@siemens.com, nickel@dbs.ifi.lmu.de
%}

\maketitle

\begin{abstract}

We provide a survey on relational models. Relational models   describe complete networked {domains  by taking into account global dependencies in the data}.  Relational models can lead to more accurate predictions if compared to non-relational machine learning approaches. Relational models typically are based on probabilistic graphical models, e.g., Bayesian networks, Markov networks, or latent variable models. Relational models  have applications in  social networks analysis, the modeling of  knowledge graphs, bioinformatics, recommendation  systems, natural language processing, medical decision support,   and  linked data.

\end{abstract}

\section{Synonyms}
 Relational learning,
 statistical relational models,
 statistical relational learning,
 relational data mining

% TERMS (DO NOT ELETE)
% vertices =  nodes
% arcs =  directed edges =  arrows
% tie
% edge
% link
%  SNA: individuals, actors, nodes ... nodes (representing individual actors within the network)  ... ties (which represent relationships between the individuals
% Markov: nodes, edges
% Bayes net: nodes, directed edges
% RDF: nodes and labeled directed arcs (link: linked data)
% dyadic relations are usually called binary relations

% relation name;
% relation instance = actual table
% tuple
% tuple \in relation instance
% predicate is a function (specific to world)
% predicate(tuple) is true or false

\section{Glossary}

\begin{description}

   \item[\textbf{Entities}]  are (abstract) objects. We denote an entity by a lowercase $e$. An actor in a social network can be modelled as an entity.
    There can be multiple types of entities in a domain  (e.g., individuals, cities, companies), entity attributes (e.g., income, gender) and relationships between entities (e.g., knows, likes, brother, sister).
   Entities, relationships and attributes are defined in the entity-relationship model, which is used in the design of  a formal relational model

 \item[\textbf{Relation}] A {relation or relation instance $I(R)$} is a  set of tuples. A tuple $t$  is an ordered list of elements $(e_1, e_2, \ldots, e_{\textit{arity}})$, which, in the context of this discussion, represent entities. {The \textit{arity} of a relation is the number of elements in each of its tuples, e.g., a relation might be unary, binary or higher order. }
      $R$ is the name or type of the relation. For example, \textit{(Jack, Mary)} might be a tuple of the relation instance \textit{knows}, indicating that Jack knows Mary.
     A database instance (or world) is a {set of relation instances. } {For example, a database instance  might contain
      instances of the unary relations     \textit{student, teacher, male, female},  and instances of the
      binary relations \textit{knows, likes, brother, sister}} (see Figure~\ref{fig:rdb})

 \item  [{\textbf{Predicate}}] A {predicate $R$} is a mapping of tuples to true or false. $R(t)$ is a \textit{ground} predicate and is true when $t \in I(R)$, otherwise it is false. Note that we do not distinguish between the relation name $R$ and the predicate name $R$.  Example: \textit{knows} is a predicate and \textit{knows(Jack, Mary)} returns \textit{True} if  it is true that Jack knows Mary, i.e., that $\textit{(Jack, Mary) }\in I(\textit{knows})$. The convention is that relations and predicates are written in lowercase and entities in uppercase

 \item[\textbf{Probabilistic Database}]
A (possible) world corresponds to a database instance.
   In a probabilistic database,   a probability distribution is defined over all possible worlds under consideration. Probabilistic databases with potentially complex  dependencies can be  described by  probabilistic graphical models.
In a canonical representation,  one assigns a binary random  variable $X_{R, t}$ to each possible tuple $t$ in each relation $R$. Then
\[
t \in I(R) \;  \Leftrightarrow \; R(t)= \textit{True} \; \Leftrightarrow  \;  X_{R, t} = 1
\]
and
\[
t  \notin I(R) \;  \Leftrightarrow  \; R(t)= \textit{False} \; \Leftrightarrow \;  X_{R, t} = 0   .
\]
The probability for a world $x$ is written as  $P( X = x)$
where $X = \{ X_{R, t} \}_{R, t}$ is the set of random variables and $x$ denotes their values in the world (see Figure~\ref{fig:rdb})

\item[\textbf{Triple Database}]  A triple database consists of  binary relations represented as  subject-predicate-object triples. An example of a triple is:  \textit{(Jack, knows, Mary)}. A triple database can be represented as a knowledge graph with entities as nodes and predicates as directed links, pointing  from the subject node to the object node.  The
      Resource Description Framework (RDF)  is triple based and is the basic data model of the Semantic Web's Linked Open Data.
      In social network analysis, nodes would be individuals or actors and links would correspond to ties

  \item [\textbf{Linked Data}]  Linked Open Data describes a method for publishing structured data so that it can be interlinked and can be exploited by machines.
        Linked Open Data uses the RDF data model

  \item[\textbf{Collective learning}] {refers to the effect that
  an entity's relationships, attributes or class membership can be predicted  not only from  its attributes  but also from { its (social)}  network environment
}

  \item[\textbf{Collective classification}] A special case of collective learning: The class membership of entities can be predicted from the class memberships  of entities in their {(social)} network environment. Example: {Individuals}' income classes  can be predicted from those of  their friends

  \item[\textbf{Relationship prediction}] The prediction of  the existence of a relationship between entities,  for example friendship between {individuals}. A relationship is typically modelled as a binary relation

  \item[\textbf{Entity resolution}] The task of predicting if  two constants refer to the {same} entity

    \item[\textbf{Homophily}] The tendency of {an individual} to associate  with similar others

  \item[\textbf{Graphical models}] A graphical description of a probabilistic domain where nodes represent random variables and edges represent direct probabilistic dependencies

  \item[\textbf{Latent Variables}] Latent variables are quantities which are not measured directly and whose states are inferred from data

\end{description}

\section{Definition}

 {Relational models} are machine-learning models that are able to truthfully {represent}  some or all distinguishing features of a relational domain such as  long-range dependencies over multiple relationships.  Typical examples for relational domains include social networks and knowledge bases.   Relational models concern  nontrivial relational domains  with at least one relation with an arity of two or larger  that  describes the relationship between entities, e.g.,\textit{ knows, likes, dislikes}. In the following we will focus on  nontrivial relational domains.

\section{Introduction}

 {Social networks can be modelled  as graphs, where actors correspond to nodes  and where relationships between actors such as friendship, kinship, organizational position, or sexual relationships are represented by directed labelled links (or ties) between the respective nodes.}
{  Typical machine learning tasks would concern the prediction of
  unknown relationship instances between actors, as well as the prediction of actors's attributes and class labels.}
{ In addition, one might be interested in a  clustering of actors.}
{  To obtain best results, machine learning
  should take an actors's network environment into account. Thus two individuals might appear in the same cluster because  they have common friends.}

  \emph{Relational learning} is a branch of machine learning that is concerned {with these tasks, i.e.~to learn efficiently from data where information is represented in form of relationships between entities.}

 \emph{Relational models} are machine learning models that  truthfully model some or all distinguishing features of relational data such as long-range dependencies propagated via  relational chains and homophily, i.e. the fact that  entities with similar attributes  are  neighbors  in the relationship structure.
In addition to social network analysis, relational models are used to model  {knowledge graphs,} preference networks, citation networks, and biomedical networks such as gene-disease networks or protein-protein interaction networks.
{Relational models can be used to solve   the aforementioned  machine learning tasks, i.e., classification, attribute prediction, clustering.}  Moreover, relational models can be used to solve {additional relational learning tasks  such as   relationship prediction and entity resolution.}
{Relational models are derived from  directed and undirected graphical models or  latent variable models and typically  define a probability distribution over a relational domain.}

\section{Key Points}

Statistical relational learning is a subfield of machine learning.  Relational models learn a probabilistic model of a complete networked {domain  by taking into account global dependencies in the data}.  Relational models can lead to more accurate predictions if compared to non-relational machine learning approaches. Relational models typically are based on probabilistic graphical models, e.g., Bayesian networks, Markov networks, or latent variable models.

\section{Historical Background}

 {Inductive logic programming (ILP)   was  maybe  the first machine learning effort  that  seriously focussed on a relational representation.} It gained attention in the early  1990s and focusses on learning deterministic or close-to-deterministic dependencies, {with representations derived from   first order logic.}
  As a field, ILP was {introduced in a} seminal paper by  Muggleton~\cite{Muggleton91}. A very early and still very influential algorithm is Quinlan's FOIL~\cite{DBLP:journals/ml/Quinlan90}. ILP will not be a focus in the following, since social networks exhibit primarily statistical dependencies.
 Statistical relational learning started around  the beginning of the millennium  with the work by
 Koller, Pfeffer, Getoor and Friedman~\cite{Koller:89,DBLP:conf/ijcai/FriedmanGKP99}.
 Since then many combinations of  ILP and relational learning have been explored. The Semantic Web, Linked Open Data are producing vast quantities of relational data and~\cite{Tresp:09,Nickel2012} describe the
application of  statistical relational learning to these emerging fields.
{Relational learning has been applied to the learning of knowledge graphs, which model large domains as triple databases. \cite{nickel2016review} is a recent review on the application of relational learning to knowledge graphs. An interesting application is the semi-automatic completion  of knowledge graphs by analysing information from the Web and other sources,  in combination with relational learning, which  exploits the information already present on the knowledge graph~\cite{dong2014knowledge}.}

\section{Machine Learning in  Relational Domains}

\subsection{Relational Domains}

Relational domains are domains that can truthfully be represented by relational databases.  The glossary defines the key terms such as a relation, a predicate,  a tuple and a database. Nontrivial relational domains contain
 at least one relation with an arity of two or larger that  describes the relationship between entities, e.g.,\textit{ knows, likes, dislikes}.  The main focus here is on   nontrivial relational domains.

Social networks are typical relational domains, where information is
{represented by  multiple types of relationships (e.g., \textit{knows, likes, dislikes}) between entities (here: actors), as well as through the attributes of entities.}

\subsection{Generative Models for a Relational Database}

%
%An  available database instance is generated by a generative probabilistic model of the form $P(X=x)$
%where $X = \{ X_{R, t} \}_{R, t}$ is the set of random variables and $x$ denotes their values (see the glossary).
%
%

{Typically, relational} models can exploit   long-range or even global dependencies   and have principled ways of dealing with missing data.
 Relational models are often displayed as  probabilistic graphical  models and  can be thought of as relational  versions of regular
graphical models, e.g., Bayesian networks, Markov networks,  and latent variable models. The approaches often have a ``Bayesian flavor'' but  a fully Bayesian statistical  treatment is  not always performed.

The following section describes common relational graphical models.

\subsection{Non-relational Learning}

Although we are mostly concerned with relational learning, it is instructive to analyse the special case of non-relational learning. Consider a database with a key entity class \textit{actor} with elements $e_i$ and with only unary relations; thus we are considering a trivial relational domain. Then one can partition the random variables into independent disjoint sets according to the entities, and  the joint distribution factorizes as
\[
 \prod_i P(\{X_{R, e_i}\}_{R})
\]
where the binary random variable  $X_{R, e_i}$  is assigned to tuple $e_i$ in unary relation $R$ (see glossary).

Thus the set of random variables can be reduced to  non-overlapping independent sets of random variables. This is the common non-relational learning setting with \textit{i.i.d.} instances, corresponding to the different actors.

\subsection{Non-relational Learning in a Relational Domain}

An common approximation to a relational model is to model unary relations of key entities in a similar way as in a non-relational model as
\[
 \prod_i P(\{X_{R, e_i}\}_{R} \;  | \; \mathbf{f}_i )
\]
where $\mathbf{f}_i$ is a vector of relational features that are derived from the relational network environment of the actor $i$. Relational features provide additional information to support learning and prediction tasks. For instance, the average income of {an individual's friends might be a good covariate to predict {an individual's} income in a social network}. {The underlying mechanism that forms these patterns might be homophily, the tendency of individuals to associate  with similar others.   The goal of this approach is to be able to use \textit{i.i.d.} machine learning by exploiting some of the relational information. This approach is commonly used  in applications where probabilistic models are computationally too expensive.
The application of  non-relational machine learning to relational domains is sometimes referred to as  \textit{propositionalization}.

 Relational features are often high-dimensional and sparse (e.g., there are many people, but only a small number of them are {an individual}'s friends; there are many items but {an individual} has only bought a small number of them) and in some domains it can be easier to define useful  kernels than to define useful features. Relational kernels often reflect the similarity of entities with regard to the network topology. For example a kernel can be defined based on counting the substructures of interest in the  intersection of two graphs defined by  neighborhoods of the  two entities~\cite{DBLP:conf/esws/LoschBR12} (see also  the discussion on RDF graphs further down).

\subsection{Learning Rule-Premises  in Inductive Logic Programming}

% Good relational features for a particular prediction task in a relational domain are not always obvious and

Some researchers apply a systematic search for good features and  consider this as an essential distinction between {\emph{relational} learning} and {\emph{non-relational}} learning:  in non-relational learning features are essentially defined prior to the training phase whereas relational learning includes a systematic and automatic search for features in the relational context of the involved entities.
  Inductive logic programming (ILP) is a form of relational learning with the goal of finding deterministic or close-to-deterministic {dependencies, which are described in logical form such as Horn clauses. Traditionally, ILP involves}  a  systematic search for sensible relational features that form the rule premises~\cite{Dz:07}.

\begin{figure}
\center
  \includegraphics[width= 1.0 \textwidth,angle=0]{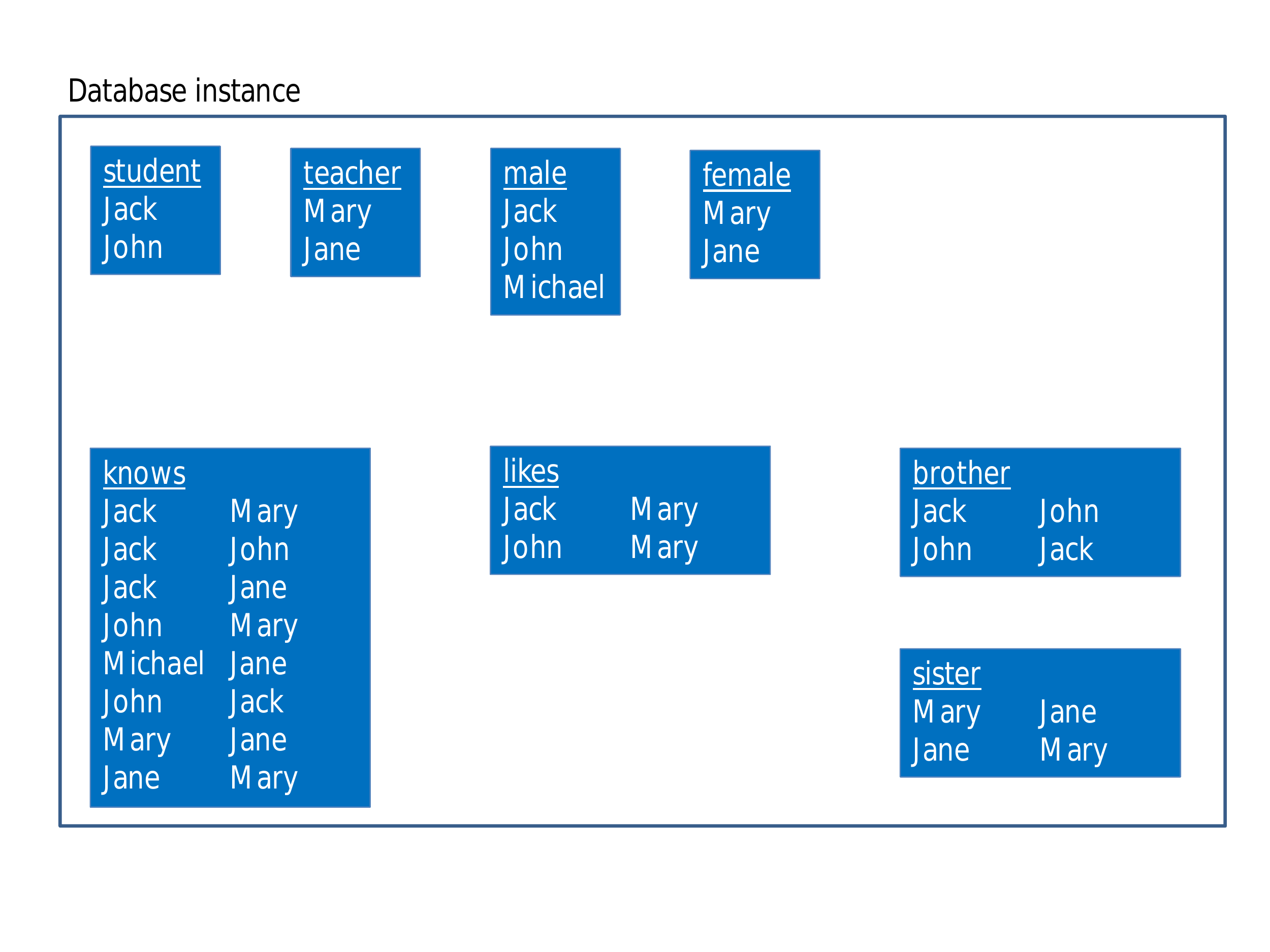} %
\caption{A database instance (world) with  4 unary relations \textit{(student, teacher, male, female)} and 4 binary relations \textit{(knows, likes, brother, sister)}.  As examples: $(\textit{Jack, Mary}) \in I(\textit{knows})$. Thus $\textit{knows(Jack, Mary)=True}$, and $X_{knows, (\textit{Jack, Mary})}=1$.  $(\textit{Jack, Michael}) \notin I(\textit{knows})$. Thus $\textit{knows(Jack, Michael)}=\textit{False}$, and $X_{\textit{knows, (Jack, Michel)}}=0$.}
\label{fig:rdb}
\end{figure}

\section{Relational Models}

In this section we describe the most important relational models in some detail. These are based on probabilistic graphical models, which efficiently model high-dimensional probability distributions by exploiting independencies between random variables}. In particular, we consider Bayesian networks, Markov networks and latent variable models.  We start with a more detailed discussion on possible world models for relational domains and with a discussion on the dual structures of
the  triple graph and the
probabilistic graph.

\subsection{Random Variables  for Relational Models}

As mentioned before,   a probabilistic database defines a  probability distribution  over the possible worlds under consideration. The goal of relational learning is to derive a model of this probability distribution.

In a canonical representation,  we assign a binary random  variable $X_{R, t}$ to each possible tuple in each relation. Then
\[
t \in I(R) \;  \Leftrightarrow \; R(t)= \textit{True} \; \Leftrightarrow  \;  X_{R, t} = 1
\]
and
\[
t  \notin I(R) \;  \Leftrightarrow  \; R(t)= \textit{False} \; \Leftrightarrow \;  X_{R, t} = 0   .
\]
The probability for a world $x$ is written as  $P( X = x)$
where $X = \{ X_{R, t} \}_{R, t}$ is the set of random variables and $x$ denotes their values in the world (see Figure~\ref{fig:rdb}).
What we have just described corresponds to a closed-world assumption where all tuples, which are  not part of  the database instance,  map to   $R(t)= \textit{False}$ and thus $X_{R, t} = 0$. In contrast  in an open world assumption, we would consider the corresponding truth values and states as being unknown and the database instance as being only partially observed. Often in machine learning some form of a \textit{local} closed-world assumption is applied with a mixture of true,  false and unknown ground predicates~\cite{dong2014knowledge,krompass2015type}. For example one might assume that, if at least one child of an individual is specified, it implies that all children are specified (closed-world), whereas if no child is specified, children are considered unknown (open-world). Another aspect is that type constraint imply that certain ground predicates are false. For example, only individuals can get married, but neither cities or buildings. Other types of background knowledge might materialize tuples that are not explicitly specified. For example, if individuals live in Munich, by simple reasoning one can conclude that they also live in Bavaria and Germany.
 The corresponding  tuples can be added to the database.

Based on background knowledge, one might want to modify the canonical representation, which uses  only binary random variables.
  For example, {discrete random variables with $N$ states are} often  used to {implement the constraint that} exactly one out {off} $N$ ground predicates is true, e.g. that {an individual} belongs exactly to one out of $N$ income classes or age classes. It is also possible to extend the model towards continuous variables.

So far we have considered an underlying probabilistic model and an observed world. In probabilistic databases one often assumes a noise process between the actual database instance and the observed database instance by specifying a conditional probability
\[
P(Y_{R, t} | X_{R, t}) .
\]
Thus only $Y_{R, t}$ is observed whereas the real interest is on $X_{R, t}$:
One observes a $t \in I^y(R) \;  \Leftrightarrow  \;  Y_{R, t} = 1$ from which one can infer for the database instance
$P(t \in I(R)) \;  \Leftrightarrow  \;  P(X_{R, t} = 1)$. With an observed  $Y_{R, t} = 1$,  there is a certain probability that $X_{R, t} = 0$ (error in the database) and with an observed
$Y_{R, t} = 0$ there is a certain probability that $X_{R, t} = 1$ (missing tuples).

 The theory of probabilistic databases focussed on the issues of complex query answering under a probabilistic model. In probabilistic databases~\cite{DBLP:series/synthesis/2011Suciu} the canonical representation is used in tuple-independent {databases, while multi-state}  random variables are used in block-independent-disjoint (BID) databases.

Most relational models assume that all entities (or constants) and all predicates are known and fixed (domain closure assumption).  In general these constraints can be relaxed, for example if one needs to include new individuals in the model. Also, latent variables derived from a cluster or a  factor analysis can be interpreted as new ``invented'' predicates.

\subsection{Triple Graphs and Probabilistic Graphical Networks}

\begin{figure}
\center
  \includegraphics[width= 1.0 \textwidth,angle=0]{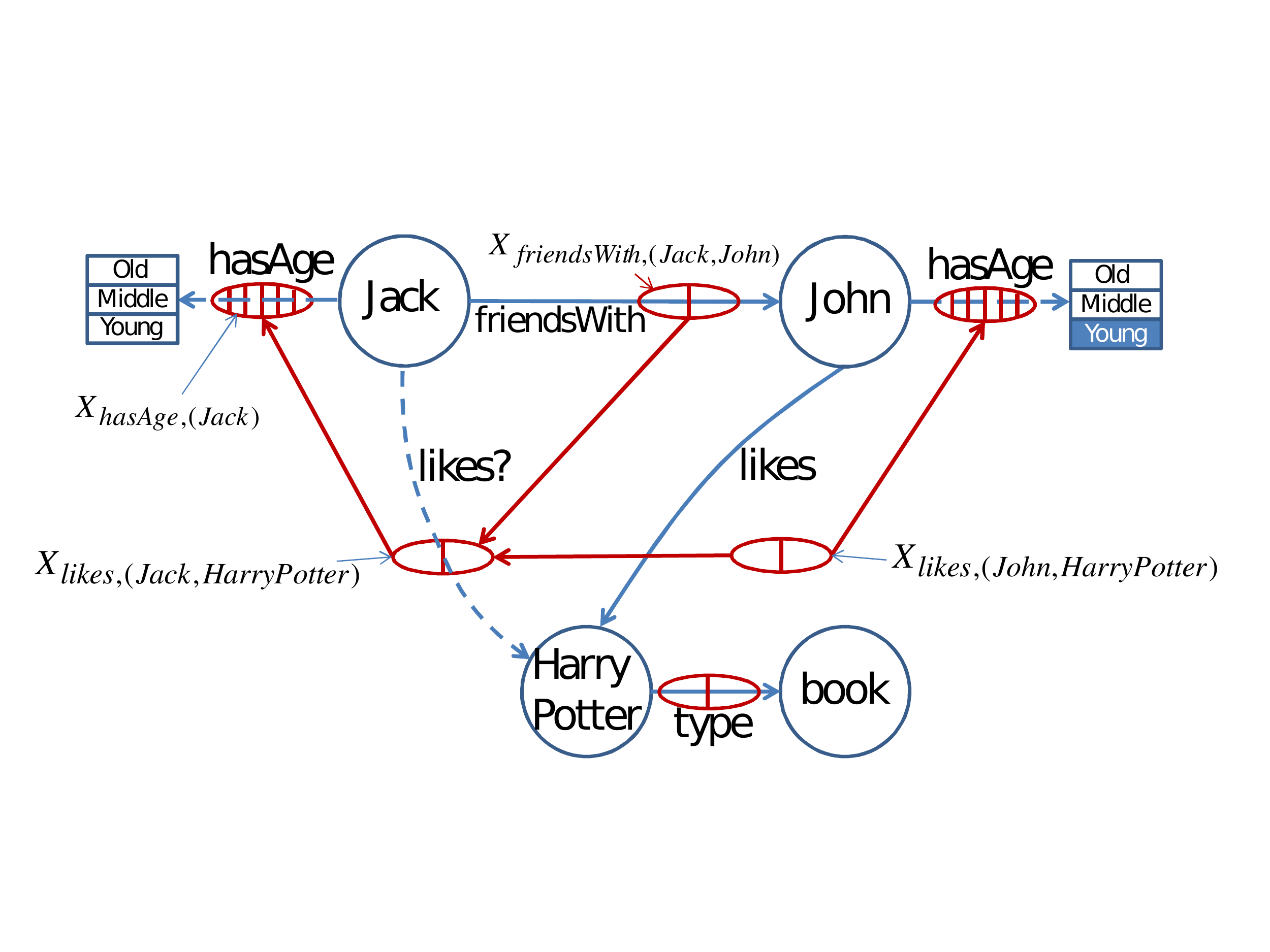} %
\caption{The figure clarifies the relationship between the triple graph and the probabilistic graphical network.
 The round nodes stand for entities in the domain, square nodes stand for attributes,   and the labelled links stand for triples.  Thus we assume that it is known that \emph{Jack} is friends with \emph{John} and that \emph{John} likes the book
 \emph{HarryPotter}.  The oval nodes stand for random variables and their states represent the existence (value 1) of non-existence (value 0) of a given labelled link; see for example the node $X_{likes, (John,  HarryPotter)}$ which represents  the ground predicate  \emph{likes(John,  HarryPotter)}. Striped oval nodes stand for random variables with many states, useful for attribute nodes (exactly one out of many ground predicates is true). The unary relations \textit{hasAgeOld, hasAgeMiddle,} and \textit{hasAgeYoung} are represented by the random variable $X_{hasAge}$ which has three states. Relational models assume a probabilistic dependency between the probabilistic nodes. So the relational model might learn that \emph{Jack} also likes \emph{HarryPotter} since his friend \emph{Jack} likes it (homophily). Also $X_{likes, (John,  HarryPotter)}$ might correlate with the  age of John.  The direct dependencies are indicated by the red  edges between the elliptical nodes. In PRMs the edges are  directed (as shown)  and in Markov logic networks they are undirected. The elliptical random nodes and their quantified edges form a probabilistic graphical model. Note that the probabilistic  network is dual to the triple  graph in the sense that links in the triple graph become nodes in the probabilistic  network.
}
\label{fig:rdf}
\end{figure}

 A triple database consists of  binary relations represented as  subject-predicate-object triples. An example of a triple is:  \textit{(Jack, knows, Mary)}. A triple database can be represented as a knowledge graph with entities as nodes and predicates as directed links, pointing  from the subject node to the object node.
  Triple databases are able to represent web-scale   knowledge bases and  sociograms that allow multiple types of directed links.  Relations of higher order can be reduced to binary relations by introducing auxiliary entities (``blank nodes'').  Figure~\ref{fig:rdf}  shows an example of a triple graph.
  The
      Resource Description Framework (RDF)  is triple based and is the basic data model of the Semantic Web's Linked Open Data.
      In social network analysis, nodes would be individuals or actors and links would correspond to ties.

For each triple a random variable is introduced.
  In {Figure~\ref{fig:rdf}} these random variables are represented as elliptical  red nodes.
The binary random variable associated with the tripe $(s=i, p=k, o=j)$ will be denoted as $X_{k(i,j)}$.

\subsection{Directed Relational Models}
\label{sec:DRGM}

The  probability distribution of a directed relational model, {i.e.~a}  relational
Bayesian model, can be written as
\begin{equation}
P \left(\{ X_{R, t} \}_{R, t} \right) = \prod_{R, t} P(X_{R, t} | \textit{par}(X_{R, t})) .
\label{eq:drm}
\end{equation}
Here { $\{ X_{R, t} \}_{R, t}$ refers to the set of random variables  in the directed relational model, while $X_{R, t}$ denotes a particular random variable}. In a graphical representation, directed arcs are
pointing from all parent nodes $\textit{par}(X_{R, t})$ to the node $X_{R, t} $ (Figure~\ref{fig:rdf}). As {Equation~\ref{eq:drm} indicates the model requires the specification of the parents of a node and the specification of the probabilistic dependency of a node, given the states of  its parent nodes. In specifying the former,  one often follows a causal ordering of the nodes, i.e., one assumes that the parent nodes causally influence  child nodes and their descendents. An important constraint is that the resulting directed graph is not permitted to have directed {loops, i.e.~that it is a directed acyclic graph}. A major challenge is to specify  $P(X_{R, t} | \textit{par}(X_{R, t}))$, which might require the calculation of  complex aggregational features as intermediate steps.

\subsubsection{Probabilistic Relational Models}
% \label{sec:prm}

Probabilistic relational models (PRMs) were one of the first published  directed relational models and found great interest
in the statistical machine learning
community~\cite{Koller:89,Getoor:07}. An example of a PRM is shown in Figure~\ref{fig:prm}. PRMs combine a frame-based (i.e., object-oriented)
logical representation with probabilistic semantics based on
directed graphical models.  The PRM provides a template for specifying the graphical probabilistic structure and the quantification of the probabilistic dependencies for any ground PRM.
In the basic PRM models   only the entities' attributes are uncertain whereas the relationships between
entities are {assumed to be known}.  Naturally, this assumption  greatly simplifies
the model.  Subsequently, PRMs have been extended to also consider the case that
relationships between entities are unknown, which is
called \textit{structural uncertainty} in the PRM
framework~\cite{Getoor:07}.

In PRMs one can distinguish parameter learning and structural learning.
 In the simplest case the dependency structure is known and the truth values of all ground predicates are known as well in the training data. {In this case,  parameter}  learning consists of estimating parameters in the conditional probabilities.  If the dependency structure is unknown,
structural
learning is applied, which  optimizes an appropriate cost function and typically  uses a greedy search strategy to find the optimal dependency structure. In structural learning,  one needs
to guarantee that the ground Bayesian network does not contain
directed loops.

In general the data will contain missing information, i.e., not all truth values of all ground predicates are known in the available data.
 For some PRMs, regularities in the PRM structure can be exploited (encapsulation) and even exact inference to estimate the missing information
is possible. Large PRMs require approximate inference; commonly,
loopy belief propagation is being used.

\begin{figure}
\center
  \includegraphics[width= 1.0 \textwidth,angle=0]{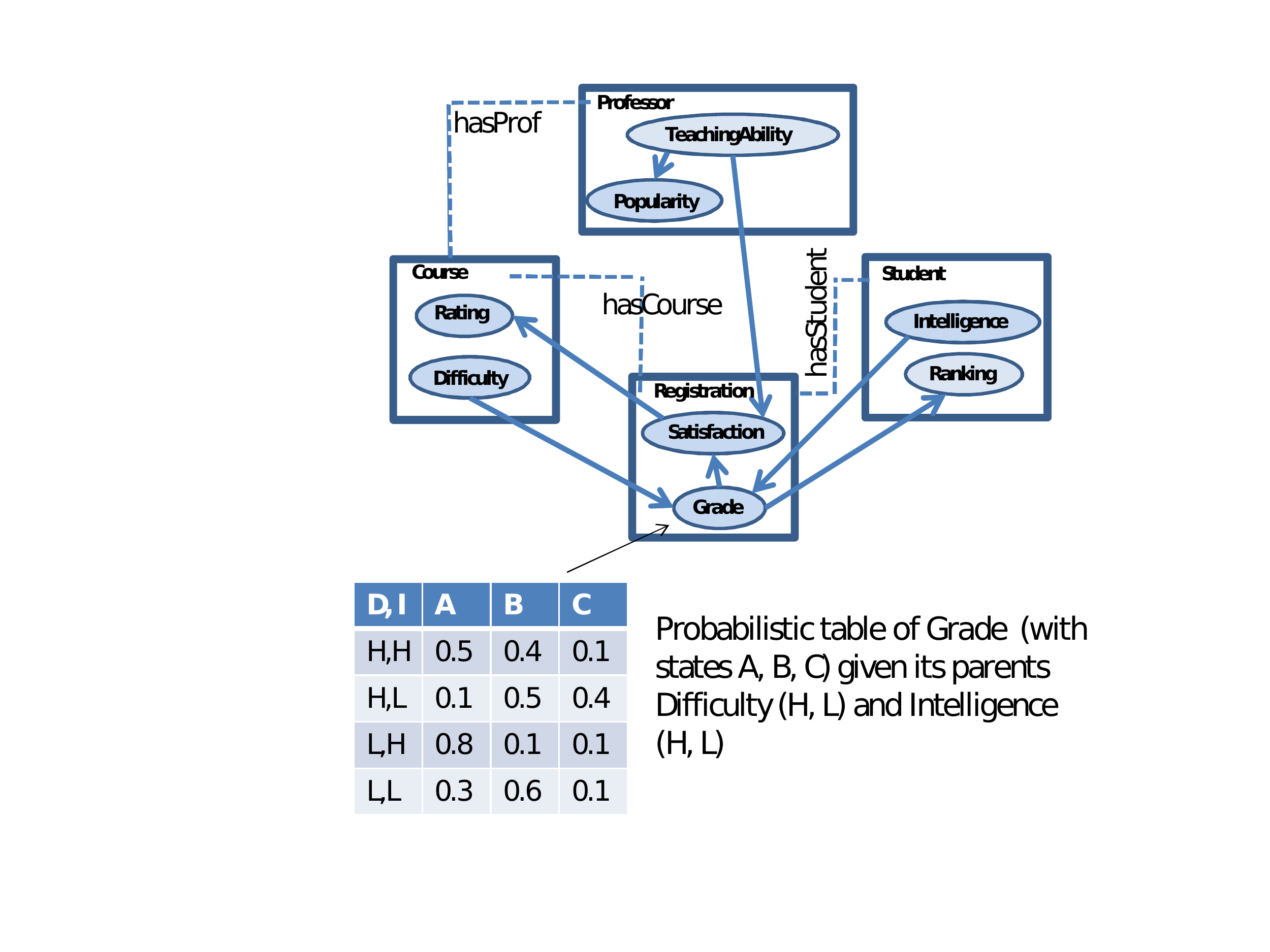} %
\caption{A PRM with domain predicates
 \emph{Professor(ProfID, TeachingAbility, Popularity)},
 \emph{Course(CourseID, ProfID, Rating, Difficulty)},
 \emph{Student(StuID, Intelligence, Ranking)}, and
 \emph{Registration(RegID, CourseID, StuID,  Satisfaction, Grade)}.  Dotted lines indicate foreign keys, i.e. entities defined in another relational instance.
 The directed edges indicate direct probabilistic dependencies on the template level.
   Also shown is a probabilistic table of the random variable \emph{Grade} (with states \emph{A, B, C}) given its parents \emph{Difficulty} and \emph{Intelligence}. Note that some probabilistic dependencies work on multisets and require some form of aggregation: for example  different students might have different numbers of registrations and the ranking of a student might depend on the (aggregated) average grade from different registrations. Note the complexity in the dependency structure which can involve several entities: for example the \emph{Satisfaction} of a \emph{ Registration} depends on the the \emph{TeachingAbility} of the \emph{Professor} teaching the \emph{Course} associated with  the \emph{Registration}.  Consider the additional complexity when structural uncertainty is present, e.g., if the \emph{Professor} teaching the \emph{Course} is unknown.
}
\label{fig:prm}
\end{figure}

\subsubsection{More Directed Relational Graphical Models}

A {Bayesian logic program} is defined as a set of Bayesian clauses~\cite{Kersting:01}. A Bayesian clause specifies the conditional probability distribution of  a random variable given its parents. A special feature is that, for a given random variable, \emph{several} such
conditional probability distributions might {be given and combined}  based on
various combination rules (e.g., noisy-or). In a Bayesian logic
program, for each clause there is one conditional probability
distribution and for each random variable
there is one combination rule.
\textit{Relational Bayesian
networks}~\cite{Jaeger:97} are related to Bayesian logic programs
and use probability formulae for specifying conditional
probabilities.
The probabilistic entity-relationship (PER) models~\cite{Heck:07} are related to the PRM framework and use the entity-relationship model as a basis, which is often used in the design of a relational database.
{Relational dependency networks~\cite{Neville:2004} also belong to the
 family of directed relational  models and  learn the dependency of a node given its Markov blanket (the smallest node set that make the node of interest independent of the remaining network). Relational dependency networks are generalizations of dependency networks as introduced by~\cite{Heck:00, Hof:97}.  A  relational dependency networks typically contains directed loops and thus is not a proper Bayesian network.

\subsection{Undirected Relational Graphical Models}

The probability distribution of an undirected graphical model, i.e. a Markov network, is written as a log-linear model in the form
\[
P\left(X  = x  \right) =
 \frac{1}{Z} \exp \sum_{i}  w_i f_i (x_{i})
\]
where the feature functions $f_i$ can be any real-valued function
on the set  $x_{i} \subseteq  x$ and where $w_i \in \mathbb{R}$. In a probabilistic graphical representation one forms undirected edges between  all nodes that jointly appear in a feature function.
 {Consequently, all nodes that appear jointly in a function will form a \emph{clique} in the graphical representation.}
 $Z$ is the partition function normalizing the
distribution.

A major advantage is that undirected graphical models can elegantly model symmetrical dependencies,  which are common in  social networks.

\subsubsection{Markov Logic Network (MLN)}

 A Markov logic network (MLN) is a probabilistic logic which combines Markov networks with first-order logic.
 In MLNs the random variables, representing ground predicates, are part of a Markov network, whose dependency structure is derived from a {set of first-order} logic formulae (Figure~\ref{fig:mln}).

 Formally, a MLN $L$
 is defined as follows: Let $F_i$ be a first-order formula, (i.e., {a logical  expression containing constants, variables, functions and predicates) and let $w_i \in \mathbb{R}$ be a weight attached to each formula. Then  $L$ is defined as a set of pairs
 $(F_i, w_i)$~\cite{Richardson:06, Domingos:07}.

 From $L$  the ground Markov network $M_{L, C}$   is generated as follows.  First, one generates nodes (random variables) by introducing a binary node for each possible grounding of each predicate  appearing in $L$  given a set of
constants $c_1, \ldots, c_{|C|}$ (see the discussion on the canonical probabilistic representation).
The state of a node is equal to one if the ground  predicate is true, and zero otherwise.
The feature functions $f_i$, which define the probabilistic dependencies in the Markov network,  are derived from the formulae by grounding them in a domain. For
formulae that are universally quantified,  grounding  is an assignment of constants to the variables in the
formula.
If a
formula contains $N$ variables, then there are $|C|^{N}$ such
assignments.
The  feature function  $f_i$ is equal to one if the ground formula is true,  and  zero otherwise.
The  probability distribution of the $M_{L, C}$  can then be written as
\[
P\left(X = x \right) = \frac{1}{Z}
 \exp\left(
 \sum_i w_i n_i(x)
 \right) ,
 \]
where $n_i(x)$ is the number of formula groundings that are
true for $F_i$  and where the weight $w_i$ is associated with formula
 $F_i$ in $L$.

  {The joint distribution $P\left(X = x\right)$} {will be maximized when large weights are assigned to formulae that are frequently true}.
   In fact,  the larger the weight, the higher is
the confidence that a formula is true for many groundings.
Learning in MLNs consists of estimating the weights $w_i$ from data. In learning, MLN makes a
closed-world assumption and employs a  pseudo-likelihood cost
function, which is the product of the probabilities of each node
given its Markov blanket. Optimization is performed using a limited
memory BFGS algorithm.

The simplest form of {inference in a MLN} concerns the prediction of the truth
value of a ground predicate given the truth values of other
ground {predicates. For this task an efficient algorithm can be derived:  In
	the first phase of the algorithm, the}
minimal subset of the ground Markov network
{is computed that}
 is
required to calculate the conditional {probability of the queried ground predicate}. It is essential
that this subset is small since in the worst case, inference could
involve all nodes.  In the second {phase, the conditional probability is then computed by applying Gibbs sampling to the reduced network}.

Finally, there is the issue of structural learning, which, in this
context, means the learning {of first} order formulae.
Formulae can be learned by directly optimizing the
pseudo-likelihood cost function or by  using ILP algorithms. For the
latter, the authors use CLAUDIAN~\cite{Raedt:97}, which can learn
arbitrary first-order clauses (not just Horn clauses, as in many other
ILP approaches).

An advantage of MLNs is that the features and thus the dependency structure is defined using a well-established  logical representation. On the other hand, many people are unfamiliar with logical formulae and might consider the PRM framework to be more intuitive.

\begin{figure}
\center
  \includegraphics[width= 1.0 \textwidth,angle=0]{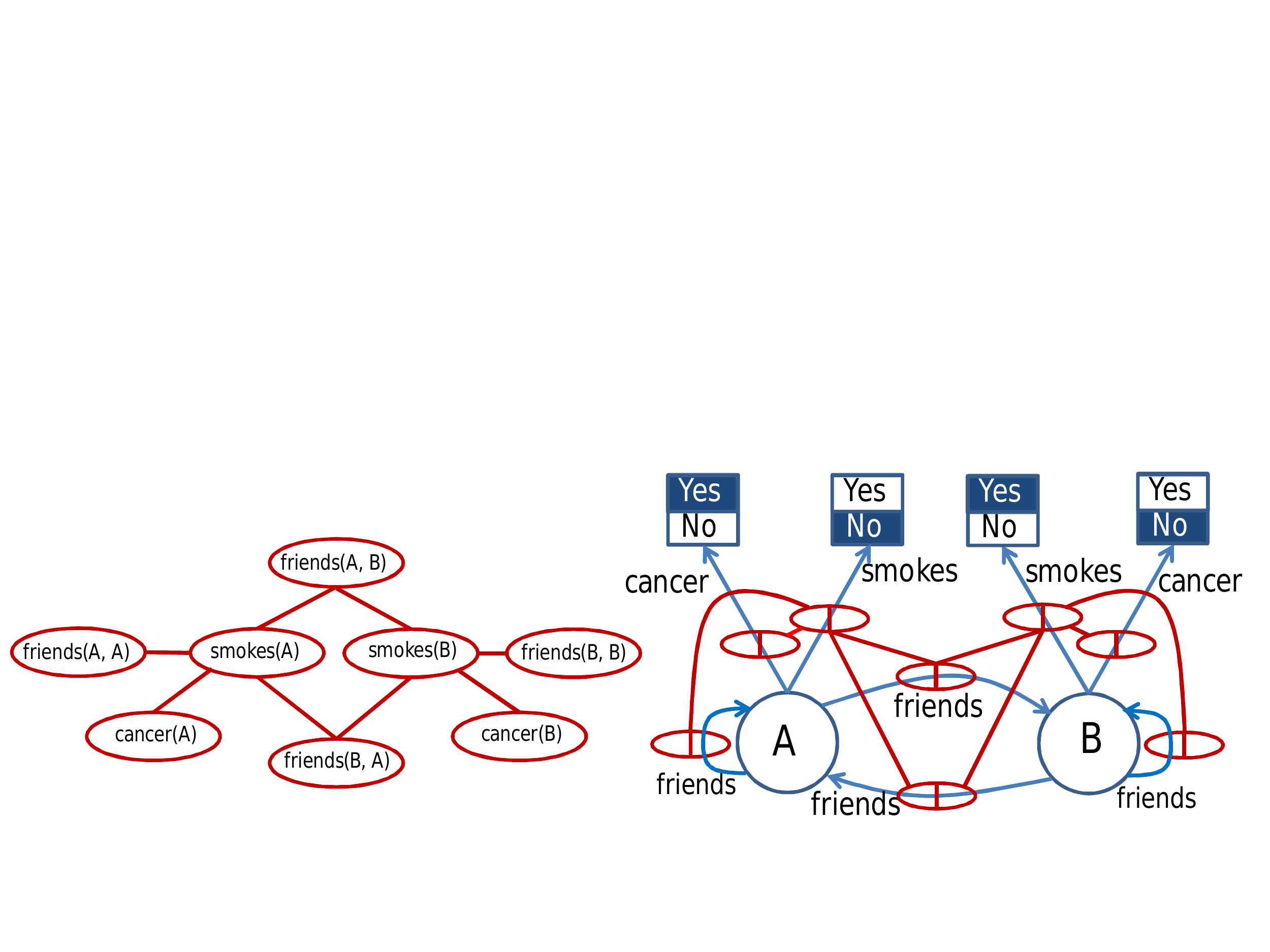} %
\caption{Left: An example of a MLN. The domain has two entities (constants) \emph{A} and \emph{B }   and the unary relations \emph{smokes} and \emph{cancer} and the binary relation \emph{friends}. The 8 elliptical nodes are the ground predicates. Then there are two logical expressions $\forall x \; smokes(x) \rightarrow cancer(x)$  (someone who smokes has cancer) and $\forall x\forall y \; friends(x, y) \rightarrow (smokes(x) \leftrightarrow smokes(y)) $ (friends either both smoke or both do not smoke).  Obviously and fortunately both expressions are not always true and learned weights on both formulae  will assume finite values.
 There are two groundings of the first formula (explaining the edges between the \emph{smokes} and \emph{cancer} nodes) and four groundings of the second formula, explaining the remaining edges. The corresponding features are equal to one if the logical expressions are true and are zero else. The weights on the features are adapted according to the actual statistics in the data. Redrawn from~\cite{Domingos:07}.
 Right: The corresponding triple graph for two individuals (blue) and the dual ground Markov network (red).
}
\label{fig:mln}
\end{figure}

\subsubsection{Relational Markov Networks (RMNs)}

RMNs generalize many concepts of PRMs to undirected
relational  models~\cite{DBLP:conf/uai/TaskerPK02}. RMNs use conjunctive database queries as
clique templates, where a clique  in an undirected graph is a subset of its nodes such that every two nodes in the subset are connected by an edge. RMNs are mostly trained discriminately. In contrast to {MLNs and similarly to PRMs,
RMNs do} not make a closed-world assumption during
learning.

\subsection{Relational  Latent Variable Models}
\label{IHRM}

In  the approaches described so far, the structures in the  graphical models were either defined using expert knowledge or were learned directly from data using some form of structural learning.
Both can be problematic since appropriate expert domain knowledge might not be {available, while structural learning can be very time consuming and possibly results} in local optima which are difficult to interpret.
{In this context, the advantage {of relational latent variable models} is} that the structure in the associated  graphical models is purely defined by the entities and relations in the domain.

  The additional complexity of working with a latent representation is counterbalanced by the great simplification by avoiding  structural learning. In the following discussion, we assume that data is in triple format;  generalizations to relational databases haven been described~\cite{Xu:06, krompass2014probabilistic}.

 \begin{figure}
\center
    \includegraphics[width= 1.0 \textwidth,angle=0]{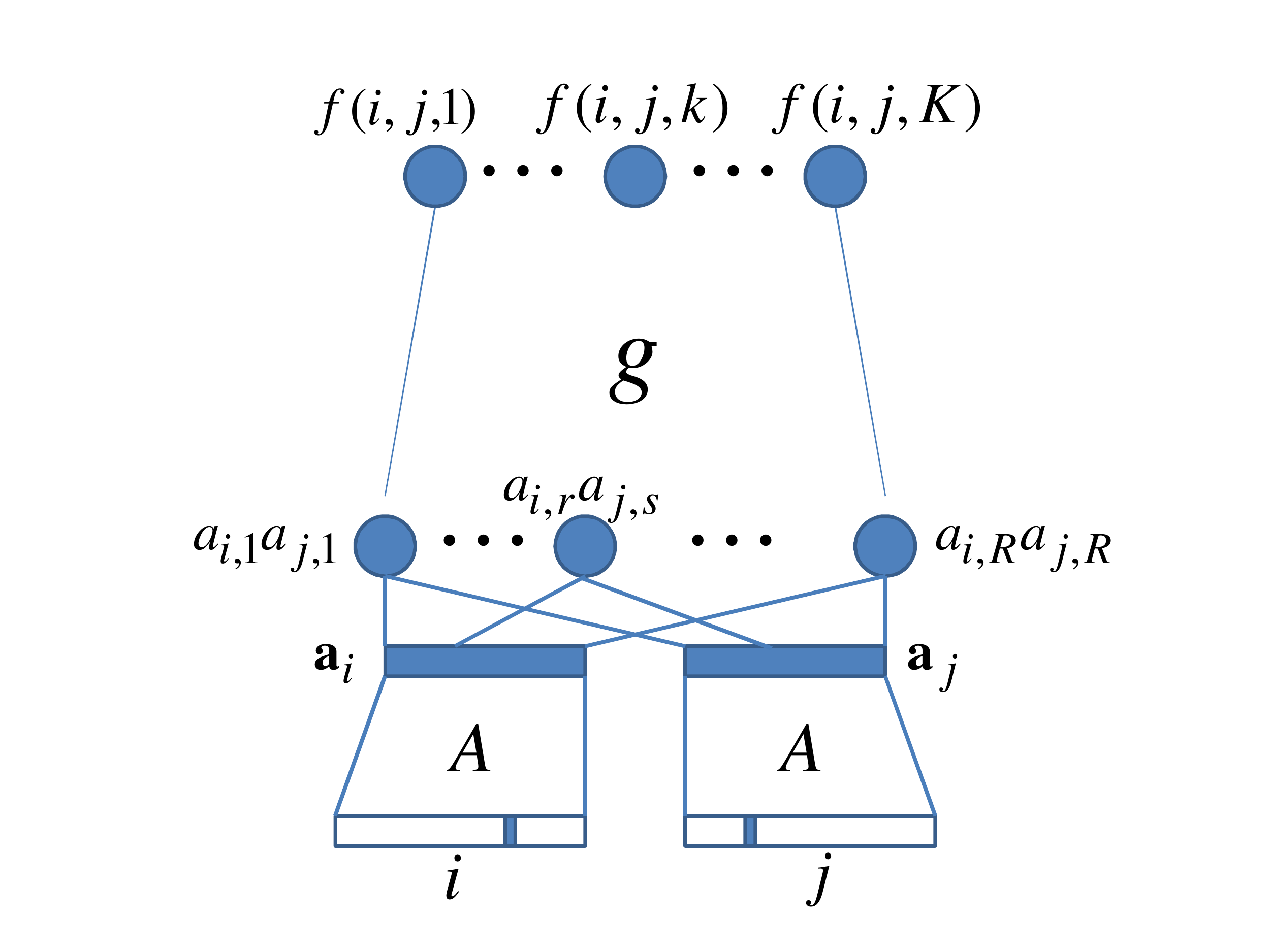} %
\caption{
The architecture of RESCAL and the IHRM. In the bottom layer (input) the index units for subject $s=i$ and object $o=j$ are activated (one-hot encoding). $A$ is a weight matrix.  The second layer  calculates the latent representations $\mathbf{a}_i$ and $\mathbf{a}_j$. The following layer forms componentwise products.
The output layer then calculates
$
f(i, j, k) = \sum_{r, q} a(i, r) a(j, q) g(k, r, q)
 $ for predicate $p=k$.
 For RESCAL,  $P(X_{k(i,j)}| i, j) = \textit{sig}(f(i, j, k))$. In the IHRM, we identify $P(r|i) = a(i, r)$,
$P(q|j) = a(j, q)$, $P(X_{k(i,j)}|r, q) = g(k, r, q)$ and
 $P(X_{k(i,j)} | i, j) = f(i, j, k)$.
For the IHRM, the factors must be nonnegative and properly normalized. $R$ is the number of latent dimension and $K$ is the number of relations.
}
\label{fig:rdflatent}
\end{figure}

\subsubsection{The IHRM: A Latent Class Model}

The  infinite hidden relational model (IHRM)~\cite{Xu:06} (a.k.a infinite relational model~\cite{Kemp:06}) is a generalization to a probabilistic mixture model where a latent variable with states $1, \ldots, R$ is assigned to each entity  $e_i$. If the latent variable for subject $s=i$ is in state $r$ and   the latent variable for object $o=j$ is in state $q$, then the triple $(s=i, p=k, o=j)$ exists with probability    $P(X_{k(i,j)}|r, q)$. Since the latent states are unobserved, we obtain
\[
P(X_{k(i,j)}| i, j) = \sum_{r, q} P(r|i) P(q|j) P( X_{k(i,j)}   |r, q)
\]
which can be implemented as the sum-product network of  Figure~\ref{fig:rdflatent}.

In the {IHRM  the number of states (latent classes) in each latent variable is allowed to be
infinite   and  fully Bayesian learning is performed based on  a    Dirichlet process mixture model. For inference  Gibbs sampling is employed where only a small number  of the infinite states are
occupied in sampling, leading to a clustering solution where the number of
states in the latent variables
  is automatically determined. Models with a finite number of states have been studied as stochastic block models~\cite{nowicki2001estimation}.

Since the dependency structure in the ground Bayesian network is
local, one might get the impression that only local information
influences prediction. This is not true, since
latent representations are shared and
in the ground
Bayesian network the latter are parents to the random network   variables $X_{k, (i, j)}$. Thus
common children  with evidence lead to
interactions between the parent latent variables. Thus information
can propagate in the network of latent variables.

The IHRM has  a number of key advantages. First, no structural
learning  is required, since the directed arcs in the ground
Bayesian network are directly given by the structure of the triple graph. Second, the IHRM model can be thought of as an infinite
relational mixture model, realizing hierarchical Bayesian modeling.
 Third, the mixture model can be used for   a cluster  analysis providing insight into  the relational domain.

The IHRM has been applied to social networks, recommender systems, for gene function
prediction and to develop medical recommender systems. The IHRM was
the first relational  model applied to trust
learning~\cite{Rettinger:08}.

In~\cite{AiroldiBFX08} the IHRM is generalized to a mixed-membership stochastic block model, where entities can belong to several classes.

\subsubsection{RESCAL: A Latent Factor Model}

The RESCAL model was introduced in~\cite{Nickel2011} and  follows a similar dependency structure as the IHRM as shown in Figure~\ref{fig:rdflatent}. The main differences are  that, first,  the latent variables do not describe entity classes but are latent entity factors and that, second,  there are no nonnegativity or normalization constraints on the factors.
The probability of a triple  is calculated with
\begin{equation}\label{eq:tens}
  f(i, j, k) = \sum_{r, q} a(i, r) a(j, q) g(k, r, q)
\end{equation}
as
\[
P(X_{k(i,j)}| i, j) = \textit{sig} \left( f(i, j, k)\right)
\]
where $\textit{sig}(x) = 1/(1 + \exp -x)$.

As in the IHRM, factors are unique to entities which leads to
interactions between the factors in  the ground
Bayesian network, enabling  the propagation of information
in the network of latent factors.
The relation-specific matrix $G^{R} = g(k, :, :)$ encodes the factor interactions  for a specific relation and its asymmetry permits the representation of directed relationships.

The calculation of {the latent factors is} based on the factorization of a multi-relational adjacency tensor where two modes represent the entities in the domain and the third mode represents the relation type (Figure~\ref{fig:RESCAL}).
With a closed-world assumption and a squared-error cost function, efficient alternating least squares (ALS) algorithm can be used; for local closed world assumptions and open world assumptions,  stochastic gradient descent is being used.

{The relational learning capabilities of the RESCAL model} have been demonstrated on classification tasks and entity resolution tasks, i.e., the mapping of entities between knowledge bases.  One of the great advantages of the RESCAL model is its scalability: RESCAL has been applied to the YAGO ontology~\cite{DBLP:conf/www/SuchanekKW07} with several million entities and 40 relation types~\cite{Nickel2012}! The YAGO ontology, closely related to DBpedia~\cite{DBLP:conf/semweb/AuerBKLCI07} and the Google Knowledge Graph~\cite{Singhal},  contains formalized knowledge from Wikipedia and other sources.

RESCAL is part of a tradition on relation prediction using factorization of matrices and tensors. \cite{YuCYTX06} describes a Gaussian process-based approach for predicting a single relation type, which has been generalized to a mutli-relational setting in~\cite{XuKT09}.

 \begin{figure}
\center
  \includegraphics[width= 1.0 \textwidth,angle=0]{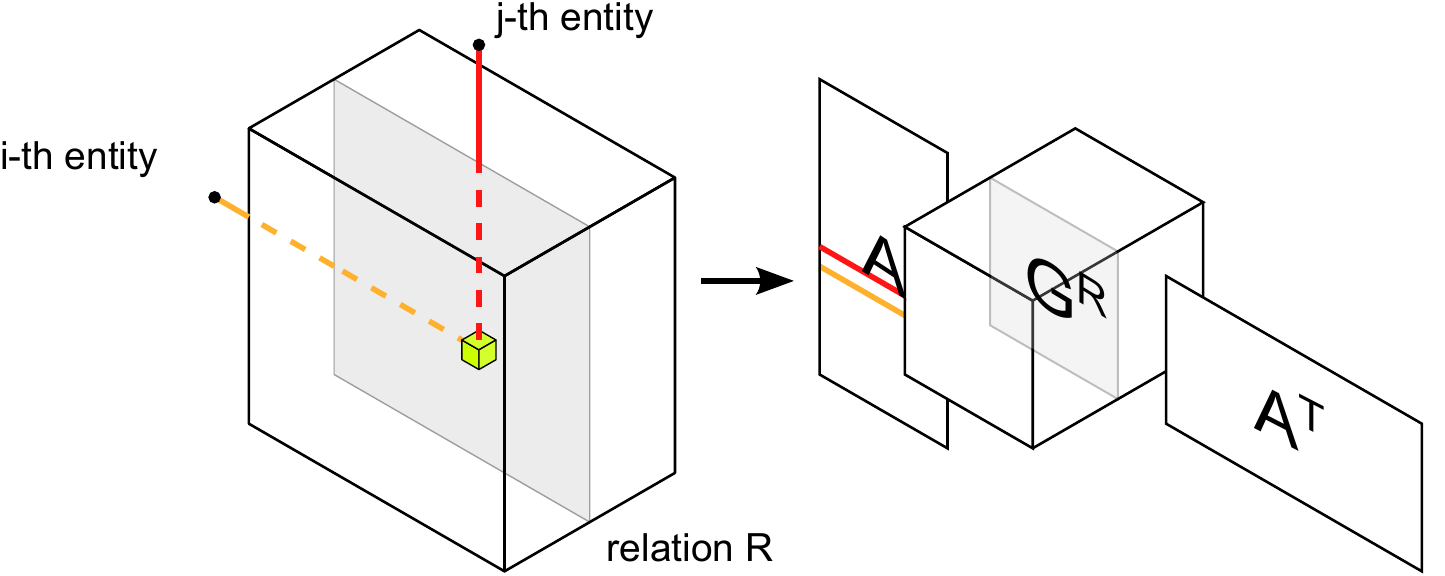} %
\caption{In RESCAL, Equation~\ref{eq:tens} describes a tensor decomposition of the tensor $\mathcal{F} = f(:, :, :)$ into the factor matrix $A= a(:, :)$ and core tensor $\mathcal{G} = g(:, :, :)$.
In the  multi-relational adjacency tensor on the left,   two modes represent the entities in the domain and the third mode represents the relation type. The $i$-th row of the matrix $A$ contains the factors of the $i$-th entity.  $G^R$ is a slice in the $\mathcal{G}$-tensor and encodes the relation-type specific factor interactions. The factorization can be interpreted as a  constrained Tucker2 decomposition.
}
\label{fig:RESCAL}
\end{figure}

A number of variations and  extensions exist. The SUNS approach~\cite{Tresp:09} is based on a Tucker1 decomposition of the adjacency tensor,  which can be computed  by a singular value decomposition (SVD). The Neural Tensor Network ~\cite{socher2013reasoning} combines several tensor decompositions. Approaches with a smaller memory footprint are  TransE~\cite{bordes2013translating} and HolE~\cite{nickel2015holographic}. The multiway neural network in the Knowledge Vault project~\cite{dong2014knowledge} combines the strengths of latent factor models and neural networks and was successfully used in semi-automatic completion of knowledge graphs. \cite{nickel2016review} is a recent review on the application of relational learning to knowledge graphs.

\section{Key Applications}

Typical applications of relational models are in  social networks analysis, knowledge graphs, bioinformatics, recommendation  systems, natural language processing, medical decision support,   and  Linked Open Data.

\section{Future Directions}

As a number of publications have shown, best results can be achieved by committee solutions integrating factorization approaches with user defined or learned rule patterns~\cite{nickel2014reducing, dong2014knowledge}. The most interesting application in recent years was in projects involving large knowledge graphs, where performance and scalability could clearly be demonstrated~\cite{dong2014knowledge,nickel2016review}.  The application of relational learning to sequential data and time series opens up new application areas, for example in clinical decision support and sensor networks~\cite{esteban2015predicting,esteban2016coev}. ~\cite{tresp2015learning} studies the relevance of relational learning to cognitive brain functions.

% BibTeX users please use
\bibliographystyle{plain}
\bibliography{RelationalModelsSpringerEnc2016}   % name your BibTeX data base

\end{document}